\documentclass[11pt]{article}

\usepackage{arxiv}

\usepackage[utf8]{inputenc} 
\usepackage[T1]{fontenc}    
\usepackage{hyperref}       
\usepackage{url}            
\usepackage{booktabs}       
\usepackage{amsfonts}       
\usepackage{nicefrac}       
\usepackage{microtype}      
\usepackage{graphicx}
\usepackage{natbib}
\usepackage{doi}

\usepackage{amsmath}
\usepackage{amssymb}
\usepackage{float}
\usepackage{enumitem} 
\usepackage{subcaption}

\usepackage{pgfplots}
\DeclareUnicodeCharacter{2212}{−}
\usepgfplotslibrary{groupplots,dateplot}
\usetikzlibrary{patterns,shapes.arrows}
\pgfplotsset{compat=newest}

\title{Impact of Underwater Image Enhancement on Feature Matching}
\date{July 29, 2025} 

\author{\href{https://orcid.org/0009-0006-4161-2865}{\includegraphics[scale=0.08]{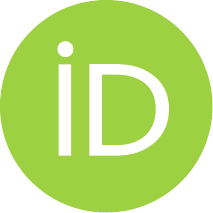}\hspace{1mm}Jason M. Summers} \\
	Department of Computer Science\\
	Swansea University\\
	Swansea \\
	\texttt{903702@Swansea.ac.uk} \\
	\And
	\href{https://orcid.org/0000-0001-8991-1190}{\includegraphics[scale=0.08]{figures/orcid.pdf}\hspace{1mm}Mark W. Jones} \\
	Department of Computer Science\\
	Swansea University\\
	Swansea \\
	\texttt{M.W.Jones@Swansea.ac.uk} \\
}

\hypersetup{
pdftitle={Impact of Underwater Image Enhancement on Feature Matching},
pdfauthor={Jason M. Summers, Mark W. Jones},
}

\begin{document}

\maketitle

\begin{abstract}
We introduce \textit{local matching stability} and \textit{furthest matchable frame} as quantitative measures for evaluating the success of underwater image enhancement. This enhancement process addresses visual degradation caused by light absorption, scattering, marine growth, and debris. Enhanced imagery plays a critical role in downstream tasks such as path detection and autonomous navigation for underwater vehicles, relying on robust feature extraction and frame matching.
To assess the impact of enhancement techniques on frame-matching performance, we propose a novel evaluation framework tailored to underwater environments. Through metric-based analysis, we identify strengths and limitations of existing approaches and pinpoint gaps in their assessment of real-world applicability. By incorporating a practical matching strategy, our framework offers a robust, context-aware benchmark for comparing enhancement methods.
Finally, we demonstrate how visual improvements affect the performance of a complete real-world algorithm --- Simultaneous Localization and Mapping (SLAM) --- reinforcing the framework’s relevance to operational underwater scenarios.
\end{abstract}

\section{Introduction}
\label{sec:intro}

\subsection{Offshore Assets}
In recent years there has been large developments of 
offshore wind farms (OWF). 
These, and other offshore structures and assets, are in greater need of inspection and monitoring compared to structures on land because structures in an underwater environment may be subject to biological growth, turbulent currents, corrosion and physical degradation, all of which may contribute to an increase in fatigue. 
Checks and maintenance for these structures present many challenges due to the depth and length of the inspections, and so, many surveys are conducted with remotely operated underwater vehicles (ROUV or ROV) or autonomous underwater vehicles (AUV). These vehicles record hours of footage using a range of sensors, particularly optical. This footage is analysed to detect structural damage, either through environment mapping or by identifying specific regions of interest within the footage. Due to the length of the footage, it is useful to move towards the introduction of automatic analysis algorithms. 

Current image enhancement research typically evaluates performance using known or synthetically generated ground truths, along with metrics such as PSNR, SSIM, and LPIPS. However, little attention has been given to its impact on downstream tasks like feature matching, camera motion estimation, and trajectory reconstruction. In this work, we address this gap by evaluating enhancement methods through their influence on these downstream applications.

\subsection{Challenges of Underwater Data}\label{sec:intronoise}
Image enhancement is used to address several challenges presented by underwater images:

\paragraph{Challenge 1 - Occlusion and Illumination:}

Inconsistencies in visibility can arise from the varying depths, capture angles, sedimentation and properties of the water which can greatly affect a model's ability to classify elements of an underwater image \citep{jalal2020fish}. Objects such as bubbles, debris and biological life can often block regions of the footage, potentially causing the inspector to miss key parts of a structure. Light degradation can also be an issue, the Lambert-Beer empirical law states that decay in the intensity of light depends on the properties of the medium through which the light travels, so the water itself can alter the colour and illumination of objects \citep{raveendran2021underwater} \citep{schettini2010underwater}. For every 10 metres of depth underwater, the light available is halved \citep{zhang2019survey}. Visibility can also vary with weather, storms can create turbulent water, creating a more complex medium that amplifies the issues above. Light degradation and absorption are not constant for all wavelengths of light, longer wavelengths like red and orange are more easily absorbed in contrast to shorter wavelengths like blue and green. Light from the surface is therefore altered with depth where footage often has a blue-green tint.

\paragraph{Challenge 2 - Noise and Distractions:}

Underwater scenes frequently encounter various sources of noise. Biological debris, known as marine snow, can cause challenges with backlighting, while live fish and bubbles are prominent elements that can affect deep vision models significantly. Additionally, video and image capture may introduce blurring, video noise, and lens distortion. These factors collectively pose barriers to effective image analysis tasks, including feature or object identification, for both humans and vision models. For vision models being used by the inspector, artifacts can negatively affect the contrast of images, or lead to poor embedding spaces or inaccurate similarity scores in the case of deep models. This challenge is particularly pronounced in video inspections, where ROV footage often spans multiple hours.

\paragraph{Challenge 3 - Non-stable Video Capture:}

Turbulent water causes the ROV to have unpredictable movement and video capture hindering the temporal value that video could provide. The stability of footage is an important factor in accessing the trajectory of objects in the footage \citep{shadrin2017experimental} which in turn, can affect the feature and object detection capabilities of ML models \citep{shruthi2019path} \citep{ess2009moving}.

\paragraph{Objectives:} We set out to answer the question of how we measure enhancement with regard to real-world applications using the following objectives:

\begin{enumerate}
    \item Identify and analyse the current approaches for image enhancement in the underwater domain.
    \item Examine the current metrics used to quantify and compare underwater image enhancements.
    \item Propose new measures of feature matching consistency over multiple subsequent frames and furthest frame matching to evaluate enhancement.
    \item Incorporate these into a new framework for evaluating the real-world applicability of underwater image enhancement.
\end{enumerate}

\section{Previous Research}
\label{sec:literature}


In this section, we review underwater image enhancement including how to improve an image, and how to assess the quality of the improvement.
We report classical approaches (Section \ref{sec:litclassical}), followed by a look into deep learning-based methods (Section \ref{sec:litdeeplearning}) and cover the methods of quality assessment and image evaluation (Section \ref{sec:litevaluation}).

\subsection{Classical Approaches}
\label{sec:litclassical}

Despite the rise in interest in deep learning in recent years, we found that non-deep learning-based methods still play a significant role in underwater image enhancement.

Popular methods are histogram equalisation (HE), modifying channel priors, and wavelet transforms. These methods share a common approach of spectral augmentation, in that they manipulate regions across the light spectrum to improve contrast. However, while histogram equalization generally provides enhancement across the entire spectrum, wavelet transforms and channel priors focus on specific regions with more discrimination. 
Histogram manipulation and equalisation was a focus of many publications \citep{bai2020underwater} \citep{li2020hybrid} \citep{luo2021underwater} \citep{park2021sand} \citep{jin2022color} \citep{peng2022underwater} \citep{xiang2023research}. A basic form of this is global HE, which applies this method to an entire image using a transformation function derived from the image cumulative distribution function (CDF). Histogram equalisation methods are often described in this field as having limited effectiveness in underwater scenes due to the nature of light degradation \citep{peng2022underwater} \citep{song2020enhancement}. Consequently, HE techniques have evolved to address domain specific image characteristics, including resulting from the physical properties of water.

The most common improvement in the literature is to perform HE at different scales to limit the influence of noise. Such methods include Contrast Limited Adaptive Histogram Equalization (CLAHE) \citep{reza2004realization}, an important variant designed to enhance contrast in noisy images. Although much of its early use was in medical imagery, it has garnered interest in the underwater imagery domain, being used as a baseline for multiple experiments in the literature \citep{xiang2023research} \citep{zhang2022marine}. It works by performing HE on tiles of an image while limiting the contrast to prevent over-amplification of noise, before being interpolated to form the enhanced image.

\citep{bai2020underwater} adopt another multi-scale approach, applying both global Histogram Equalization (GHE) and local Histogram Equalization (LHE) to address different aspects of enhancement. 
They then adopt a fusion strategy inspired by \citep{ancuti2017color} to combine the luminance, saliency, and exposure weight maps generated from the equalized components into the final enhanced image.

The idea of using fusion to combine various representations is a popular one across both classical and deep methods \citep{lu2017underwater} \citep{bai2020underwater} \citep{li2020low} \citep{lin2023underwater} \citep{qian2023underwater} \citep{liu2019mlfcgan} \citep{yang2021laffnet} \citep{tian2022underwater} \citep{jiang2023automatic}. \citep{ancuti2017color} present a multi-scale fusion pipeline that breaks the image into Laplacian and Gaussian pyramids which are blended at each level. Saliency mapping is considered to improve the visibility of objects in the scene. The outcome demonstrates an effective enhancement technique that offers advantages for industrial computer vision. Severely hazy and unevenly illuminated scenes, including industrial scenes, all sourced from a large selection of cameras, were among those considered during their evaluation.

\subsection{Deep Learning Based Methods}
\label{sec:litdeeplearning}
Deep model training continues to be one of the biggest areas of research in this field, growing rapidly. 
Some relatively older designs continue to remain relevant but with updated features and techniques being incorporated over time.

The challenge of image enhancement, using deep learning, primarily revolves around the use of image-to-image models. These models typically comprise an encoder, responsible for generating a new representation of the input image, and a decoder, tasked with reconstructing or enhancing the image based on this new representation. These are broadly referred to as encoder-decoder models but many variations exist.

Convolutional Neural Networks (CNN) are widely adopted for feature extraction, using convolving filters or kernels to capture spatial information within image data. Since their inception, CNNs have been prominent in much of the computer vision field, and remain so for underwater image enhancement, being heavily used in many architectures in the literature \citep{zhou2023surroundnet} \citep{jiang2023underwater} \citep{wang2021uiec} \citep{liu2019mlfcgan} \citep{li2019underwater}. Attention mechanisms are a more recent addition to the machine learning arsenal, with their efficacy notably demonstrated in 2017 by \citep{vaswani2017attention}. In the context of images, a spatial attention mechanism can be employed to prioritize and focus on different spatial regions in an image, according to their relevance to the task at hand. Their full potential is still being realised but we can already see examples in the literature that utilise this mechanism for underwater enhancement \citep{wang2020gan} \citep{qi2022sguie} \citep{siddiqua2023macgan} \citep{zhong2023scauie}.

Concerning overall architectural frameworks, it is clear from the literature that generative adversarial networks (GANs) had a major impact and influence on the field. 
Although GANs are adept at generating realistic data, their general functionality does not inherently facilitate image enhancement tasks. CycleGAN \citep{zhu2017unpaired} is a design that learns mappings between two different domains to facilitate unpaired image-to-image translation tasks and utilises encoder-decoder feature extraction much like an autoencoder. Multiple studies \citep{liu2019mlfcgan} \citep{wang2020gan} \citep{panetta2021comprehensive} \citep{xu2022image} \citep{cong2023pugan} \citep{qian2023drgan} \citep{siddiqua2023macgan} \citep{jiang2023underwater} build on the idea of GANs by using the discriminator to critique the reconstruction capabilities of an autoencoder. Namely, it will discriminate between undistorted images from the source dataset and distorted images that have been enhanced via the autoencoder. 

FUnIE-GAN \citep{islam2020fast} is a successful example of the utilisation of a multitude of methods. They present a convolutional encoder-decoder system that uses an adversarial loss generated by a discriminator for perceptual image enhancement. The design of the encoder-decoder is a U-Net \citep{ronneberger2015u} where skip-connections are used between the mirrored encoder-decoder layers. Alongside a new dataset, they also formulate an ensemble of loss functions to access perpetual image quality on numerous levels. Namely they use $L_{1}$ loss to access global similarity, a VGG-19 backed content loss function \citep{simonyan2014very} \citep{ignatov2017dslr} \citep{johnson2016perceptual}, and finally an adversarial loss.

\subsection{Evaluating Performance and SLAM}
\label{sec:litevaluation}
Objective quantitative metrics are required for performance evaluation, rather than depending on visual inspection of enhancement results. Peak Signal to Noise Ratio (PSNR) and Structural Similarity Index Measure (SSIM) dominate the literature. Other general image quality metrics included Natural Image Quality Evaluator (NIQE) \citep{mittal2012making} used by papers such as \citep{zhou2023surroundnet} \citep{sharma2023wavelength} \citep{saleem2023non} \citep{siddiqua2023macgan} \citep{wang2023domain}, Visual Information Fidelity (VIF) \citep{sheikh2006image}, used in \citep{zhou2023surroundnet} \citep{sharma2023wavelength}, and Perceptual Image Quality Evaluator (PIQE) \citep{venkatanath2015blind} used in \citep{zhou2023surroundnet} \citep{saleem2023non} \citep{siddiqua2023macgan}. There were also metrics focused specifically on underwater image quality, such as Underwater Image Quality Measure (UIQM) \citep{panetta2015human} which saw use in \citep{sharma2023wavelength} \citep{qian2023drgan} \citep{wang2023domain} \citep{han2021deep}, and Underwater Colour Image Quality Evaluation (UCIQE) \citep{yang2015underwater} which was used in \citep{yan2023hybrur} \citep{qian2023drgan} \citep{wang2023domain} \citep{han2021deep}. These metrics deliver a quick and objective analysis for the quality of an image but lack real-world implications, particularly the effect of model based image enhancement on downstream tasks. 

The process of feature matching is crucial to mapping out environments and structures using optical data, an important procedure for industrial surveys, particularly in underwater environments where structural damage is common, and inspections are needed frequently. While some studies, such as those by \citep{ancuti2017color} and \citep{yan2023hybrur}, have tested enhanced data using local feature matching between images, they have been limited to example pairs of images, rather than testing across multiple frames of a video. This ignores the importance of frame-by-frame relationships and does not represent how the enhancement impacts video-based tasks. There is a lack of literature that conducts a comprehensive comparative evaluation in this fashion and is something we address in this work. \citep{hidalgo2020evaluation} examine the behaviours of feature detection, and matching, in frames sourced from varied noisy underwater ROV videos. They compare SIFT \citep{lowe2004distinctive}, SURF \citep{bay2006surf}, ORB \citep{rublee2011orb}, BRISK \citep{leutenegger2011brisk}, and AKAZE \citep{alcantarilla2013fast} feature detection methods by recording the average features found by each detector, and the average number of inliers using nearest neighbours and homography between two consecutive frames. In the number of features detected and the number of inliers matched, both ORB and BRISK feature detectors performed the best, with SIFT performance following close behind. They additionally apply two image enhancement algorithms, a fusion filter \citep{ancuti2012enhancing}, and a backscatter removal filter \citep{zhang2016removing} to each dataset. They found that the enhancements improved the number of detected features across the board but resulted in very low improvements in inliers for all detectors except AKAZE. 
We evaluate the impact of further state-of-the-art underwater image enhancement on feature detection over longer duration.

\paragraph{SLAM:} Feature matching is a core component of Simultaneous Localization And Mapping (SLAM), for mapping 3D points and path of capture source such as a ROV in an environment. \citep{zhang2022marine} utilised sequential frame matching and SLAM performance as an empirical assessment of a model. They test CLAHE, Median Filtering (MF), and Dark Channel Prior (DCP) using ORB-SLAM 2, an implementation of SLAM that identifies 
ORB features \citep{rublee2011orb}. They 
provide both a practical metric for comparing models and guidance for optimising SLAM performance. This approach  warrants further exploration and refinement. Hence in this paper, we address the problem on two levels: first, by measuring the impact of visual enhancement on sequential frame matching ability, and second, by observing its effect on the complete pipeline, in our case, ORB-SLAM 3.

\section{Methodology}
\label{sec:method}

To find the impact of visual enhancement on downstream tasks, we apply a range of enhancement methods from the literature, including deep model methods, on underwater videos and measure their impact on feature matching. 

\subsection{Frame Matching}
\label{sec:frame_matching}

Frame matching involves identifying and correlating features between two images using feature descriptors. Location differences of matched features can indicate movement of the camera or a scene object. By comparing multiple matched features with similar displacements, the camera's relative movement can be deduced geometrically using a homography model. A higher number of features that consistently align with the model increases confidence in the estimation of the camera's motion. Matched features whose displacements do not align with the model, either due to the object motion, general noise, or occlusion, are ignored. This is commonly achieved using the Random Sample Consensus (RANSAC) algorithm, which iteratively fits the model to subsets of matches and identifies the largest set of inliers, rejecting outliers as inconsistent matches. The sensitivity of this filter is defined by a threshold for the maximum permissible distance (in pixels) between the actual position of a feature and the predicted position by the model during tuning.

\begin{figure}[ht]
    \centering
    \includegraphics[width=\textwidth]{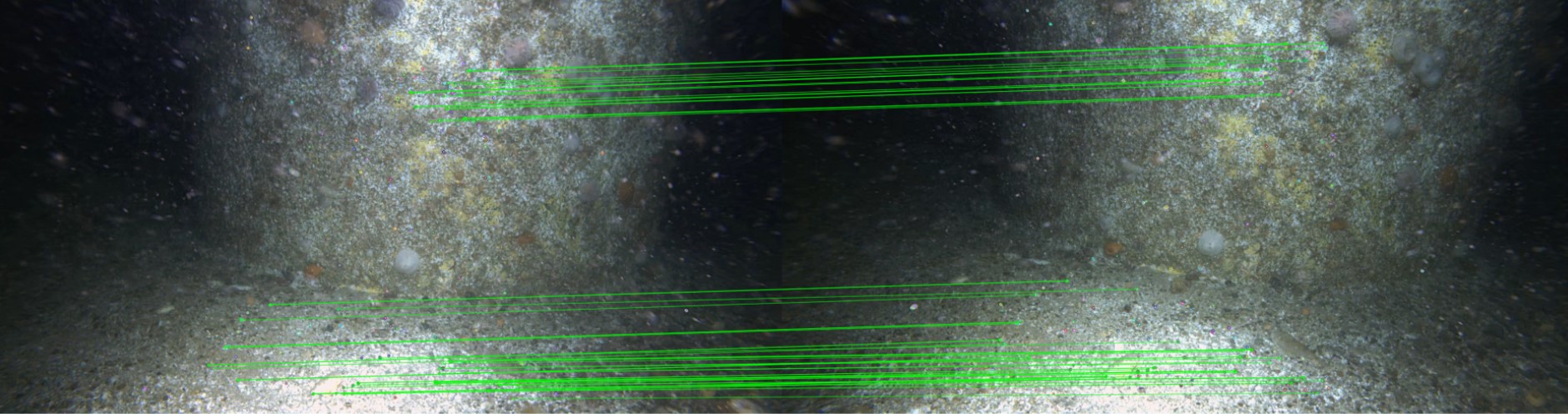}
    \caption{An example of matched features showing ideal feature tracing from two close frames in the `Seabed' video.}
    \label{fig:trace}
\end{figure}

In order to test if an enhancement better facilitates frame matching in a video, we employ two functions; One tests how a given frame in the video matches with the next $n$ consecutive frames, and the second will find the furthest frame that can still be matched given a set of threshold parameters. The threshold parameters for this study were chosen after experimentation, balancing the number of matched features to avoid both over and under-fitting. The parameters provide a suitable filter for rigorous testing while still enabling successful matches that visibly show aligned, often parallel, feature trace lines during testing, as seen in Figure \ref{fig:trace}.

\subsubsection{Feature Extraction}

The main approaches for extracting features are ORB \citep{rublee2011orb}, KAZE \citep{alcantarilla2012kaze}, AKAZE \citep{alcantarilla2013fast}, and BRISK \citep{leutenegger2011brisk},
 being available in the OpenCV library \citep{opencv_library}. SIFT \citep{lowe2004distinctive}, SURF \citep{bay2006surf} are also important types of feature extraction but have limited availability, particularly SURF, within OpenCV. Superpoint \citep{detone2018superpoint} is a keypoint detection system built using a self-supervised CNN framework, and is part of a modern wave of new deep model designs for feature detection and homography \citep{sarlin2020superglue}. 

We examine the feature methods SIFT, ORB, BRISK, KAZE, AKAZE, SuperPoint, omitting SURF due to licensing issues within OpenCV. In order to find sufficient features but maintain an acceptable processing time we limit the number of extracted features for ORB and SIFT to 1000, the number of octaves to sample for BRISK features to $4$, and set the threshold for KAZE and AKAZE to $0.001$.

\subsubsection{Local Matching Stability}

First we apply a feature finding technique on every frame in the chosen video. After extracting all features for all frames we iterate frame-by-frame and attempt to match features with the subject $frame$ with those from $frame+1$, then $frame+2$, and up to $frame+n$ for the chosen $n$, recording the camera homography data, namely the average reprojection errors for those features, inlier number, and inlier percentage. We record the matching performance over the $n=10$ frames from the subject frame.

\subsubsection{Furthest Matchable Frame Metric}

In order to compare where enhancement has affected long term feature matching, we use a similar process to our local matching function, but continue as far as possible given a threshold on the RANSAC filtered matching results. The parameters used for this threshold are as follows:
\texttt{RANSAC Threshold}: $10.0$, \texttt{Inlier Ratio}: $0.3$, \texttt{Max Reprojection Error}: $20.0$.

The \texttt{inlier ratio} refers to the minimum proportion of points that must be classified as inliers for the model to be considered valid. We found that this typically low value of $30\%$ was appropriate given the noisiness of underwater data. The \texttt{max reprojection error} measures the distance (in pixels) between the actual and model predicted positions of a feature. However, it is applied to the results of the final model rather than during the RANSAC fitting process. This value is typically closer or lower than the RANSAC threshold to ensure a stricter more robust model fit. However, due to radial lens distortion, a more lenient reprojection error threshold was used. Frames satisfying these values will yield correctly tracked camera motion with high certainty.

\subsubsection{ORB-SLAM 3}
As discussed in Section \ref{sec:litevaluation}, visual SLAM is an important procedure in underwater surveys and an key example of frame matching that is used in industry tasks. ORB-SLAM 3 is one implementation that has been frequently used as a benchmark \citep{zhang2022marine} \citep{fu2021fast} and is the state-of-the-art for SLAM implementations using ORB based features. We finalise our testing by observing the performance of SLAM after visual enhancement, using consistent tracking indicators and loop closures as measures.

\subsection{Test Datasets}

Our first test video is a publicly available video with loop closures and Inertial Measurement Unit (IMU) data, captured and used in a study \citep{joshi2019experimental}. This video contains an exploration of a natural cave system.  The second test video is sourced from a two-hour inspection of a wind turbine base. 
This footage was captured using a ROV as it traversed near a variety of pipe structures, including anodes. The footage exhibits many of the highlighted noise forms in Section \ref{sec:intronoise}, and additional challenges, such as segments of dark and featureless backgrounds. As a result of these additional challenges, selecting a candidate segment for this test proved to be a difficult task in itself, segments of quick movement and reduced features prevented consistent completion of the ORB-SLAM 3 process. The desired video should remain challenging, be discriminating amongst techniques and still allow a complete SLAM run to be achievable. We chose a five-minute segment that starts near the seabed and continues up the wind turbine structure, referring to this video as `Seabed'.

\begin{figure}[ht]
    \centering
    \begin{subfigure}[t]{0.49\textwidth}
        \centering
        \includegraphics[height=0.175\textheight]{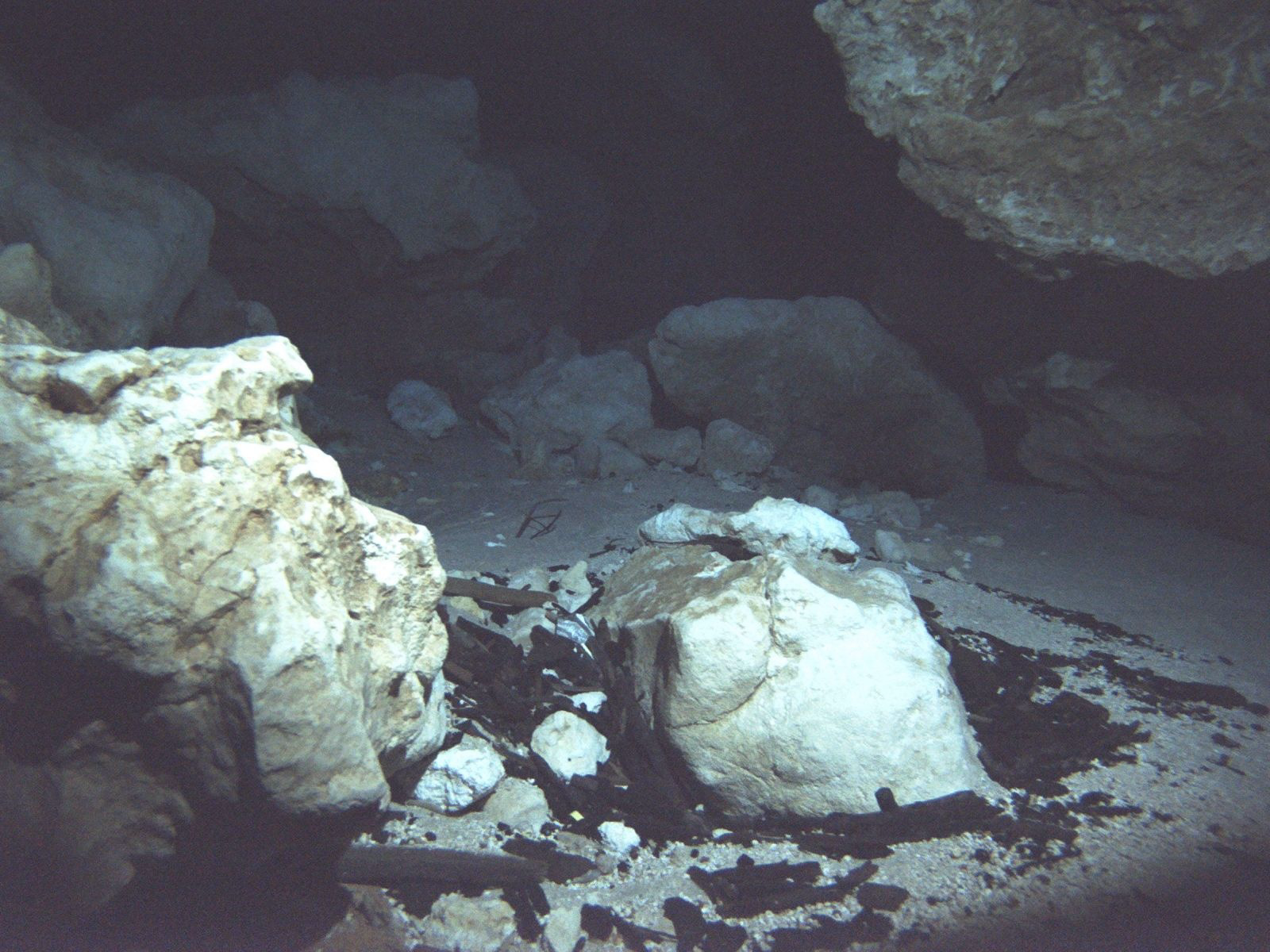}
        \caption{`Cave'}
    \end{subfigure}
    \hfill
    \begin{subfigure}[t]{0.49\textwidth}
        \centering
        \includegraphics[height=0.175\textheight]{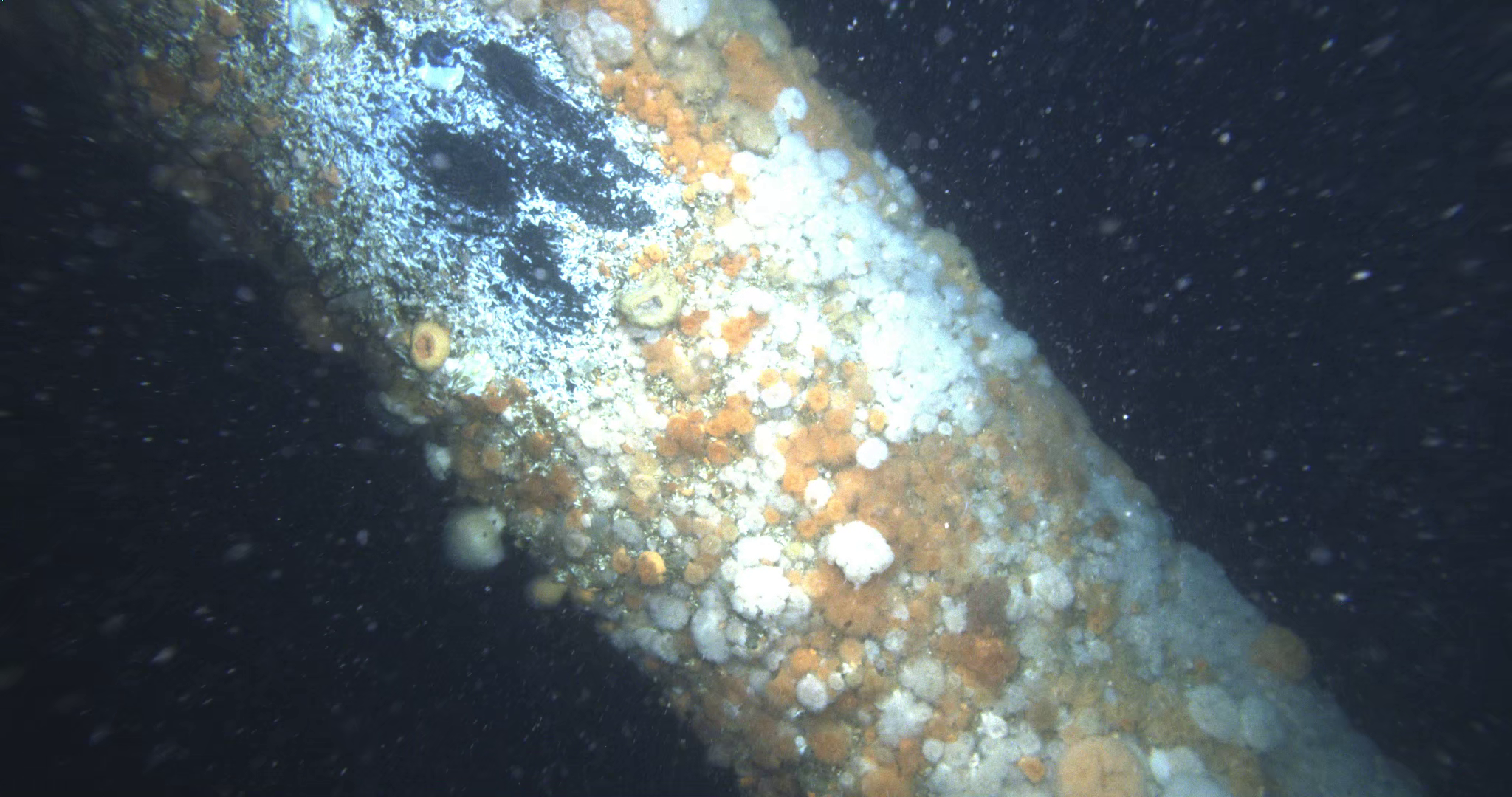}
        \caption{`Seabed'}
    \end{subfigure}
    \caption{Sample frames from `Cave' and `Seabed' videos.}
\end{figure}

\subsection{Test Models}

We test models developed by \citep{ancuti2017color}, FUnIE GAN \citep{islam2020fast}, WaveNet \citep{sharma2023wavelength}, \citep{demir2023low}, WaterNet \citep{li2019underwater}, and finally UVENet \citep{du2024end}. This selection includes a variety of approaches, ranging from an important milestone (\citeauthor{ancuti2017color}) to the state-of-the-art (\citeauthor{sharma2023wavelength} \citep{du2024end}), and covers both classical approaches \citep{ancuti2017color} \citep{demir2023low} and deep models \citep{li2019underwater} \citep{islam2020fast} \citep{sharma2023wavelength} \citep{du2024end}. 
  
Each enhancement method is individually applied to each frame of the videos to create the test video set. UVENet \citep{du2024end} highlighted a challenge found among other recent models, particularly those with temporal elements. These models can struggle to process long, high-definition videos. This limitation stems in part from their architectural design and from the limited sizes, both in length and definition, of their training clips. In order to not encounter memory issues while maximising quality, the `cave' video was divided into three parts and individually enhanced. Even so, subsequent upscaling was needed to obtain a definition matching the original.

\section{Results}
\label{sec:results}

\subsection{Classic Enhancement Validation}

An initial analysis was conducted using classic metrics, namely PSNR and SSIM, to provide a baseline for our results and ensure the enhancement pipeline was functioning correctly.

\begin{table}[ht]
\centering
\begin{tabular}{lcc}
\toprule
Method & SSIM (\(\uparrow\)) & PSNR (\(\uparrow\))\\
\midrule
FUnIE-GAN \cite{islam2020fast}         & 0.8753            & 21.6752           \\
\cite{ancuti2017color}                & 0.8394            & 24.1483           \\
\cite{demir2023low}                   & 0.8585            & 22.8294           \\
WaterNet \cite{li2019underwater}      & \textbf{0.9209}   & \textbf{24.8468}  \\
WaveNet \cite{sharma2023wavelength}   & 0.7132            & 20.7234           \\
UVENet \cite{du2024end}                  & 0.6287            & 17.4586           \\
\bottomrule
\end{tabular}
\caption{The SSIM and PSNR averages across the Seabed video. Higher is better for both SSIM and PSNR.}
\label{tab:psnr_scores}
\end{table}

These values, particularly the SSIM scores, are comparable to those from the original research papers, and overall suggest successful enhancements by each of those methods on these challenging new datasets. 
Based on these values, Waternet \citep{li2019underwater} performs the best.

\subsection{Frame Matching Benchmark}

After running our measures discussed in Section \ref{sec:frame_matching}, we have for each subject frame, a furthest matchable frame (FMF) value, and the number of inliers, inlier percentage and the average reprojection error for the next ten frames after each subject frame.

\subsubsection{Inlier Decay}

Figure \ref{fig:inlier_cave} shows the mean number of inliers for the next ten frames, across the whole video for each enhancement method using the ORB detector. The decay curve demonstrates that the next frame (1) has the highest number of inlier matches, and each subsequent frame has fewer matches with the subject frame compared to the previous frame.
The video enhanced by WaterNet, WaveNet and the unaltered original video perform consistently the best, with WaveNet and the original video being almost tied in Figure \ref{fig:inlier_cave}. The video enhanced by \citeauthor{demir2023low} using a sharpening-smoothing image filter with a CLAHE, performs the poorest in facilitating inlier detection.

\begin{figure}[h]
    \centering
    \resizebox{0.6\textwidth}{!}{
\begin{tikzpicture}

\definecolor{chocolate213940}{RGB}{213,94,0}
\definecolor{cornflowerblue86180233}{RGB}{86,180,233}
\definecolor{darkcyan1115178}{RGB}{1,115,178}
\definecolor{darkcyan2158115}{RGB}{2,158,115}
\definecolor{darkgray176}{RGB}{176,176,176}
\definecolor{gold23622551}{RGB}{236,225,51}
\definecolor{lightgray204}{RGB}{204,204,204}
\definecolor{lime2625526}{RGB}{26,255,26}
\definecolor{orchid204120188}{RGB}{204,120,188}

\begin{axis}[
legend cell align={left},
legend style={
  fill opacity=0.8,
  draw opacity=1,
  text opacity=1,
  at={(0.03,0.03)},
  anchor=south west,
  draw=lightgray204
},
tick align=outside,
tick pos=left,
x grid style={darkgray176},
xlabel={Frame Offset},
xmin=0.55, xmax=10.45,
xtick style={color=black},
y grid style={darkgray176},
ylabel={Average Number of Inliers},
ymin=152.991948590381, ymax=455.984668325041,
ytick style={color=black}
]
\addplot [thick, chocolate213940, mark=*, mark size=2, mark options={solid}]
table {%
1 406.484742951907
2 370.152902155887
3 348.370315091211
4 330.849751243781
5 313.802819237148
6 297.825373134328
7 281.345273631841
8 266.191376451078
9 250.635986733002
10 235.745273631841
};
\addlegendentry{\tiny{Ancuti et al. 2017}}
\addplot [thick, cornflowerblue86180233, mark=*, mark size=2, mark options={solid}]
table {%
1 328.034328358209
2 294.692868988391
3 275.076119402985
4 257.641127694859
5 240.951575456053
6 225.565671641791
7 210.737479270315
8 195.75903814262
9 180.223548922056
10 166.764344941957
};
\addlegendentry{\tiny{Demir et al. 2023}}
\addplot [thick, gold23622551, mark=*, mark size=2, mark options={solid}]
table {%
1 369.971973466003
2 340.060696517413
3 321.032669983416
4 304.225538971808
5 287.295190713101
6 270.579436152571
7 254.423714759536
8 238.806467661692
9 222.386069651741
10 206.695024875622
};
\addlegendentry{\tiny{FUnIE-GAN}}
\addplot [thick, darkcyan1115178, mark=*, mark size=2, mark options={solid}]
table {%
1 442.212271973466
2 406.623548922056
3 383.941791044776
4 364.304145936982
5 344.255555555556
6 325.530679933665
7 307.269154228856
8 289.50447761194
9 271.79087893864
10 254.330679933665
};
\addlegendentry{\tiny{Original}}
\addplot [thick, lime2625526, mark=*, mark size=2, mark options={solid}]
table {%
1 402.282255389718
2 371.909784411277
3 347.161525704809
4 326.169817578773
5 307.126036484245
6 289.487562189055
7 271.676119402985
8 254.599834162521
9 238.318905472637
10 222.04776119403
};
\addlegendentry{\tiny{UVENet}}
\addplot [thick, orchid204120188, mark=*, mark size=2, mark options={solid}]
table {%
1 440.290547263682
2 405.84328358209
3 383.17943615257
4 364.229684908789
5 344.62039800995
6 325.700331674959
7 307.731674958541
8 289.867330016584
9 272.231509121061
10 255.463184079602
};
\addlegendentry{\tiny{WaterNet}}
\addplot [thick, darkcyan2158115, mark=*, mark size=2, mark options={solid}]
table {%
1 399.12271973466
2 354.273134328358
3 326.048922056385
4 303.413764510779
5 282.189054726368
6 262.951077943615
7 244.826533996683
8 226.717247097844
9 208.813266998342
10 192.469485903814
};
\addlegendentry{\tiny{WaveNet}}
\end{axis}

\end{tikzpicture}}
    \caption{Average number of inliers across subsequent frames for ORB detector in the Seabed dataset. Note that the line for WaterNet (top performer) almost perfectly overlaps the line for the original unenhanced video.}
    \label{fig:inlier_seabed}
\end{figure}
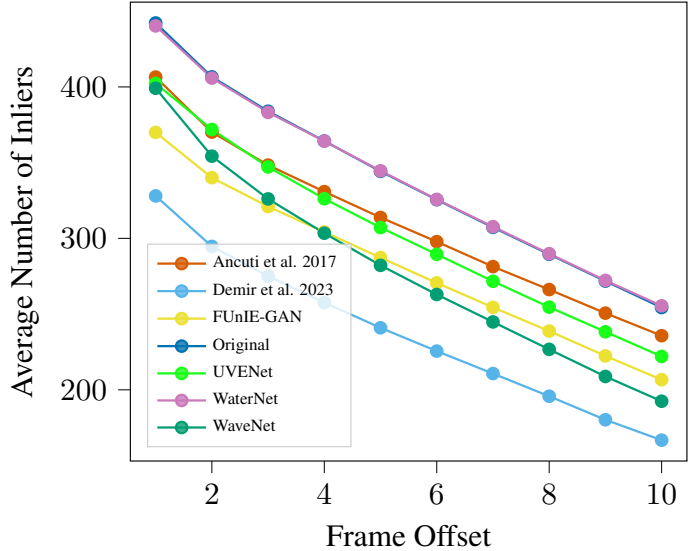

Similar results are seen in the Seabed dataset (Figure \ref{fig:inlier_seabed}) where WaterNet and the original video have almost identical results and perform the best. \citeauthor{demir2023low} is also the lowest performer. Results on the two datasets are consistent. 

\begin{figure}[h]
    \centering
    \resizebox{0.6\textwidth}{!}{
\begin{tikzpicture}

\definecolor{chocolate213940}{RGB}{213,94,0}
\definecolor{cornflowerblue86180233}{RGB}{86,180,233}
\definecolor{darkcyan1115178}{RGB}{1,115,178}
\definecolor{darkcyan2158115}{RGB}{2,158,115}
\definecolor{darkgray176}{RGB}{176,176,176}
\definecolor{gold23622551}{RGB}{236,225,51}
\definecolor{lightgray204}{RGB}{204,204,204}
\definecolor{lime2625526}{RGB}{26,255,26}
\definecolor{orchid204120188}{RGB}{204,120,188}

\begin{axis}[
legend cell align={left},
legend style={
  fill opacity=0.8,
  draw opacity=1,
  text opacity=1,
  at={(0.03,0.03)},
  anchor=south west,
  draw=lightgray204
},
tick align=outside,
tick pos=left,
x grid style={darkgray176},
xlabel={Frame Offset},
xmin=0.55, xmax=10.45,
xtick style={color=black},
y grid style={darkgray176},
ylabel={Average Number of Inliers},
ymin=117.052821202647, ymax=550.21839235076,
ytick style={color=black}
]
\addplot [thick, chocolate213940, mark=*, mark size=2, mark options={solid}]
table {%
1 481.549407402021
2 434.279211808333
3 392.996291718171
4 355.027557623791
5 319.563804260889
6 287.789791318258
7 259.666036501127
8 233.784846942485
9 211.189704064568
10 190.067839744056
};
\addlegendentry{\tiny{Ancuti et al. 2017}}
\addplot [thick, cornflowerblue86180233, mark=*, mark size=2, mark options={solid}]
table {%
1 389.421289900385
2 344.259507016651
3 308.343052424925
4 276.005671489857
5 245.8739184178
6 219.345451901403
7 195.761724714608
8 173.808187304588
9 154.437359121646
10 136.742165345743
};
\addlegendentry{\tiny{Demir et al. 2023}}
\addplot [thick, gold23622551, mark=*, mark size=2, mark options={solid}]
table {%
1 424.670108339999
2 378.529557187523
3 339.94706609467
4 304.798080418818
5 274.277103177489
6 245.923216752708
7 221.171380789646
8 198.700792554352
9 178.440776557842
10 160.051406965753
};
\addlegendentry{\tiny{FUnIE-GAN}}
\addplot [thick, darkcyan1115178, mark=*, mark size=2, mark options={solid}]
table {%
1 495.87529993456
2 450.943066967207
3 408.727768486876
4 369.223296735258
5 333.086890133062
6 300.126808696284
7 271.478659201629
8 245.499381953028
9 221.569112193703
10 199.817639787683
};
\addlegendentry{\tiny{Original}}
\addplot [thick, lime2625526, mark=*, mark size=2, mark options={solid}]
table {%
1 487.207349081365
2 427.1504311961
3 378.053768278965
4 336.582152230971
5 299.104761904762
6 265.830971128609
7 235.940982377203
8 210.309261342332
9 187.660217472816
10 167.74848143982
};
\addlegendentry{\tiny{UVENet}}
\addplot [thick, orchid204120188, mark=*, mark size=2, mark options={solid}]
table {%
1 517.970915436632
2 470.536246637097
3 427.498218570494
4 386.803824620083
5 349.243219661165
6 315.687340943794
7 285.86839235076
8 259.461208463608
9 235.032283865338
10 213.178797353305
};
\addlegendentry{\tiny{WaterNet}}
\addplot [thick, darkcyan2158115, mark=*, mark size=2, mark options={solid}]
table {%
1 530.529048207664
2 465.934559732422
3 415.476187013742
4 373.7063913328
5 335.452555806006
6 302.527739402312
7 273.400857994619
8 246.356067767033
9 222.634770595506
10 201.853341089217
};
\addlegendentry{\tiny{WaveNet}}
\end{axis}

\end{tikzpicture}}
    \caption{Average number of inliers across subsequent frames offset (1-10) from the subject frame for ORB detector in the Cave dataset.}
    \label{fig:inlier_cave}
\end{figure}
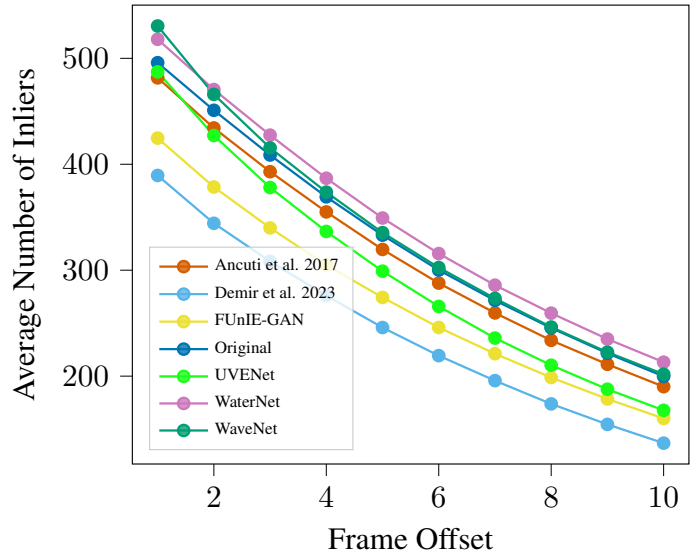

\subsubsection{Furthest Matchable Frame}

The FMF serves as a metric to evaluate how effectively a video enables optimal and consistent feature detection, showing that objects and features have the sustained clarity needed for frame matching for as long as possible. We have two approaches to evaluate the FMF values recorded for each frame. The first is the average furthest matched frame value across the whole video and is presented in Tables \ref{tab:avg_fmf_cave} and \ref{tab:avg_fmf_seabed}.


\begin{table}[h]
\centering
\begin{tabular}{lccccc}
\toprule
 & \multicolumn{5}{c}{\textbf{Detector}} \\
\cmidrule(lr){2-6}
\textbf{Enhancement Method} & \textbf{AKAZE} (\(\uparrow\)) & \textbf{BRISK} (\(\uparrow\)) & \textbf{KAZE} (\(\uparrow\)) & \textbf{ORB} (\(\uparrow\)) & \textbf{SuperPoint} (\(\uparrow\)) \\
\midrule
Ancuti et al. 2017 & 20.04 & 15.41 & 25.03 & 16.29 & 20.85 \\
Demir et al. 2023 & 15.29 & 10.76 & 20.90 & 12.57 & 22.19 \\
FUnIE-GAN & 23.98 & 19.66 & 27.09 & 14.85 & 21.28 \\
Original & 24.26 & 22.93 & 27.56 & 17.68 & 20.31 \\
UVENet & 20.78 & 22.67 & 26.59 & 15.35 & \textbf{22.23} \\
WaterNet & \textbf{24.53} & \textbf{23.58} & \textbf{29.30} & \textbf{19.25} & 20.67 \\
WaveNet & 23.86 & 22.01 & 27.63 & 17.67 & 21.10 \\
\bottomrule
\end{tabular}
\caption{Average furthest matching frame (FMF) values for the `Cave' video. A value of $k$ in the table means that on average for the whole video each frame matches with the $k^{th}$ frame ahead.}
\label{tab:avg_fmf_cave}
\end{table}


\begin{table}[h]
\centering
\begin{tabular}{lcccccc}
\toprule
& \multicolumn{6}{c}{\textbf{Detector}} \\
\cmidrule(lr){2-7}
\textbf{\small{Enhancement Method}} & \textbf{AKAZE} (\(\uparrow\)) & \textbf{BRISK} (\(\uparrow\)) & \textbf{KAZE} (\(\uparrow\)) & \textbf{ORB} (\(\uparrow\)) & \textbf{SIFT} (\(\uparrow\)) & \textbf{SuperPoint} (\(\uparrow\)) \\
\midrule
Ancuti et al. 2017 & 38.55 & 51.52 & \textbf{64.55} & 31.28 & \textbf{36.57} & 27.74 \\
Demir et al. 2023 & 29.53 & 41.75 & 43.84 & 18.78 & 20.95 & 27.51 \\
FUnIE-GAN & 38.49 & 53.08 & 54.80 & 26.41 & 26.55 & 24.52 \\
Original & 41.35 & 58.47 & 57.13 & 30.93 & 31.49 & 20.89 \\
UVENet & 39.20 & 55.19 & 56.09 & 29.33 & 24.79 & \textbf{28.52} \\
WaterNet & \textbf{42.12} & \textbf{61.80} & 60.21 & \textbf{33.04} & 32.51 & 21.82 \\
WaveNet & 36.49 & 53.12 & 49.91 & 20.17 & 23.38 & 22.69 \\
\bottomrule
\end{tabular}
\caption{Average FMF values for the `Seabed' video.}
\label{tab:avg_fmf_seabed}
\end{table}

We can see in both datasets that videos enhanced by classical methods performed poorly, with \citeauthor{demir2023low} coming last in every detector except SuperPoint, and \citeauthor{ancuti2017color} coming in the bottom three apart from a few notable exceptions. The unaltered original video for both datasets performed consistently well for all detector types, often coming second. WaveNet, UVENet, and FUnIE-GAN showed varied results, often in the middle rankings, with the one exception of SuperPoint features, where UVENet showed a good improvement over the original video. Only WaterNet showed consistently improved frame matching over all feature types, while all others were consistently outperformed by the original video.

\begin{figure}[h]
    \centering
    \resizebox{0.6\textwidth}{!}{
\begin{tikzpicture}

\definecolor{chocolate213940}{RGB}{213,94,0}
\definecolor{cornflowerblue86180233}{RGB}{86,180,233}
\definecolor{darkcyan1115178}{RGB}{1,115,178}
\definecolor{darkcyan2158115}{RGB}{2,158,115}
\definecolor{darkgray176}{RGB}{176,176,176}
\definecolor{gold23622551}{RGB}{236,225,51}
\definecolor{lightgray204}{RGB}{204,204,204}
\definecolor{lime2625526}{RGB}{26,255,26}
\definecolor{orchid204120188}{RGB}{204,120,188}

\begin{axis}[
legend cell align={left},
legend style={
  fill opacity=0.8,
  draw opacity=1,
  text opacity=1,
  at={(0.97,0.03)},
  anchor=south east,
  draw=lightgray204
},
tick align=outside,
tick pos=left,
x grid style={darkgray176},
xlabel={Frames Ahead},
xmin=-5.2, xmax=131.2,
xtick style={color=black},
y grid style={darkgray176},
ylabel={Cumulative Count},
ymin=345.9, ymax=277372.1,
ytick style={color=black},
yticklabel style={
    /pgf/number format/precision=0,
    /pgf/number format/fixed,
}, %
scaled y ticks = false,
enlargelimits = false,
]
\addplot [thick, chocolate213940]
table {%
1 13624
2 27058
3 40225
4 53060
5 65489
6 77437
7 88846
8 99671
9 109901
10 119463
11 128339
12 136536
13 144089
14 151016
15 157340
16 163117
17 168392
18 173169
19 177500
20 181446
21 185026
22 188273
23 191241
24 193930
25 196369
26 198570
27 200570
28 202366
29 203980
30 205433
31 206721
32 207873
33 208910
34 209842
35 210689
36 211483
37 212219
38 212905
39 213545
40 214148
41 214696
42 215207
43 215699
44 216158
45 216595
46 217009
47 217400
48 217779
49 218141
50 218481
51 218810
52 219122
53 219418
54 219694
55 219951
56 220194
57 220426
58 220646
59 220855
60 221053
61 221239
62 221411
63 221572
64 221724
65 221861
66 221991
67 222115
68 222235
69 222344
70 222446
71 222544
72 222638
73 222725
74 222809
75 222889
76 222968
77 223044
78 223114
79 223181
80 223244
81 223304
82 223361
83 223417
84 223472
85 223520
86 223563
87 223606
88 223645
89 223684
90 223720
91 223756
92 223787
93 223815
94 223838
95 223861
96 223879
97 223893
98 223904
99 223915
100 223925
101 223935
102 223944
103 223953
104 223960
105 223967
106 223974
107 223980
108 223985
109 223990
110 223995
111 223999
112 224002
113 224005
114 224007
115 224008
116 224009
117 224010
118 224011
119 224012
120 224013
121 224014
122 224015
123 224016
124 224017
125 224018
};
\addlegendentry{\tiny{Ancuti et al. 2017}}
\addplot [thick, cornflowerblue86180233]
table {%
1 12938
2 25202
3 36937
4 48136
5 58775
6 68820
7 78286
8 87100
9 95269
10 102761
11 109554
12 115730
13 121277
14 126250
15 130733
16 134759
17 138350
18 141552
19 144400
20 146939
21 149203
22 151217
23 153030
24 154652
25 156103
26 157383
27 158538
28 159583
29 160547
30 161439
31 162243
32 162974
33 163643
34 164255
35 164819
36 165342
37 165826
38 166281
39 166708
40 167108
41 167478
42 167817
43 168137
44 168436
45 168719
46 168979
47 169220
48 169450
49 169663
50 169864
51 170056
52 170231
53 170394
54 170544
55 170687
56 170823
57 170953
58 171070
59 171177
60 171278
61 171369
62 171459
63 171543
64 171622
65 171695
66 171764
67 171828
68 171888
69 171945
70 171996
71 172044
72 172091
73 172132
74 172172
75 172210
76 172247
77 172283
78 172318
79 172352
80 172384
81 172415
82 172445
83 172473
84 172501
85 172526
86 172550
87 172571
88 172591
89 172609
90 172627
91 172643
92 172658
93 172673
94 172688
95 172703
96 172717
97 172731
98 172744
99 172756
100 172768
101 172780
102 172791
103 172801
104 172811
105 172821
106 172829
107 172837
108 172845
109 172852
110 172858
111 172863
112 172867
113 172871
114 172874
115 172877
116 172880
117 172883
118 172884
119 172885
};
\addlegendentry{\tiny{Demir et al. 2023}}
\addplot [thick, gold23622551]
table {%
1 13322
2 26299
3 38913
4 51146
5 63000
6 74335
7 85142
8 95369
9 104972
10 113945
11 122227
12 129868
13 136870
14 143293
15 149156
16 154477
17 159274
18 163631
19 167575
20 171150
21 174403
22 177278
23 179865
24 182174
25 184236
26 186081
27 187728
28 189204
29 190514
30 191675
31 192730
32 193655
33 194495
34 195245
35 195912
36 196512
37 197039
38 197515
39 197939
40 198330
41 198679
42 199002
43 199300
44 199582
45 199843
46 200085
47 200316
48 200535
49 200746
50 200949
51 201146
52 201329
53 201504
54 201668
55 201823
56 201967
57 202104
58 202233
59 202356
60 202469
61 202575
62 202675
63 202771
64 202859
65 202938
66 203013
67 203086
68 203154
69 203219
70 203280
71 203338
72 203392
73 203440
74 203484
75 203527
76 203569
77 203609
78 203649
79 203687
80 203723
81 203759
82 203793
83 203827
84 203860
85 203892
86 203921
87 203947
88 203972
89 203997
90 204022
91 204046
92 204069
93 204092
94 204112
95 204129
96 204144
97 204156
98 204165
99 204172
100 204179
101 204185
102 204190
103 204195
104 204199
105 204202
106 204204
107 204206
108 204208
109 204210
110 204212
111 204214
112 204215
113 204216
114 204217
115 204218
};
\addlegendentry{\tiny{FUnIE-GAN}}
\addplot [thick, darkcyan1115178]
table {%
1 13509
2 26810
3 39886
4 52675
5 65162
6 77237
7 88881
8 100030
9 110593
10 120524
11 129853
12 138591
13 146775
14 154415
15 161484
16 168011
17 174082
18 179697
19 184842
20 189577
21 193879
22 197807
23 201417
24 204708
25 207710
26 210429
27 212923
28 215186
29 217263
30 219171
31 220926
32 222534
33 224008
34 225351
35 226561
36 227649
37 228638
38 229542
39 230369
40 231127
41 231830
42 232478
43 233073
44 233627
45 234146
46 234635
47 235091
48 235522
49 235935
50 236320
51 236684
52 237023
53 237348
54 237649
55 237929
56 238186
57 238438
58 238680
59 238904
60 239114
61 239317
62 239510
63 239691
64 239866
65 240037
66 240202
67 240361
68 240513
69 240661
70 240799
71 240929
72 241053
73 241175
74 241289
75 241399
76 241505
77 241603
78 241699
79 241791
80 241882
81 241967
82 242048
83 242127
84 242202
85 242272
86 242339
87 242402
88 242464
89 242522
90 242577
91 242628
92 242676
93 242721
94 242764
95 242806
96 242846
97 242883
98 242918
99 242951
100 242982
101 243011
102 243038
103 243061
104 243083
105 243103
106 243119
107 243134
108 243147
109 243160
110 243172
111 243183
112 243191
113 243199
114 243205
115 243208
116 243211
117 243213
118 243215
119 243217
};
\addlegendentry{\tiny{Original}}
\addplot [thick, lime2625526]
table {%
1 13201
2 26210
3 38938
4 51345
5 63321
6 74793
7 85669
8 95976
9 105634
10 114673
11 123060
12 130782
13 137859
14 144344
15 150254
16 155672
17 160612
18 165025
19 169025
20 172594
21 175799
22 178653
23 181194
24 183457
25 185468
26 187263
27 188870
28 190290
29 191584
30 192690
31 193655
32 194500
33 195249
34 195927
35 196533
36 197068
37 197563
38 198010
39 198409
40 198775
41 199117
42 199442
43 199753
44 200037
45 200306
46 200563
47 200799
48 201022
49 201236
50 201438
51 201628
52 201806
53 201973
54 202128
55 202277
56 202415
57 202545
58 202665
59 202777
60 202881
61 202978
62 203065
63 203144
64 203219
65 203290
66 203356
67 203420
68 203480
69 203540
70 203600
71 203657
72 203712
73 203766
74 203817
75 203867
76 203916
77 203962
78 204008
79 204053
80 204097
81 204140
82 204180
83 204220
84 204258
85 204295
86 204329
87 204361
88 204393
89 204423
90 204449
91 204475
92 204500
93 204523
94 204543
95 204563
96 204581
97 204598
98 204614
99 204627
100 204637
101 204646
102 204653
103 204659
104 204665
105 204669
106 204673
107 204675
108 204677
109 204678
110 204679
};
\addlegendentry{\tiny{UVENet}}
\addplot [thick, orchid204120188]
table {%
1 13731
2 27402
3 40964
4 54334
5 67445
6 80205
7 92557
8 104489
9 115870
10 126646
11 136767
12 146277
13 155197
14 163533
15 171347
16 178640
17 185409
18 191717
19 197565
20 202990
21 207971
22 212550
23 216718
24 220502
25 223977
26 227189
27 230128
28 232818
29 235307
30 237584
31 239670
32 241572
33 243331
34 244935
35 246400
36 247726
37 248909
38 249979
39 250959
40 251870
41 252697
42 253451
43 254148
44 254794
45 255401
46 255967
47 256492
48 256982
49 257442
50 257875
51 258277
52 258653
53 258999
54 259323
55 259624
56 259912
57 260185
58 260435
59 260673
60 260894
61 261102
62 261295
63 261481
64 261657
65 261823
66 261983
67 262132
68 262270
69 262396
70 262516
71 262632
72 262741
73 262843
74 262940
75 263036
76 263127
77 263213
78 263296
79 263377
80 263452
81 263525
82 263592
83 263656
84 263718
85 263778
86 263835
87 263892
88 263947
89 264000
90 264053
91 264106
92 264159
93 264209
94 264257
95 264302
96 264343
97 264383
98 264422
99 264460
100 264496
101 264530
102 264560
103 264589
104 264616
105 264642
106 264666
107 264687
108 264703
109 264719
110 264730
111 264740
112 264748
113 264755
114 264761
115 264765
116 264769
117 264771
118 264773
119 264775
120 264776
121 264777
122 264778
123 264779
124 264780
};
\addlegendentry{\tiny{WaterNet}}
\addplot [thick, darkcyan2158115]
table {%
1 13701
2 27303
3 40692
4 53824
5 66636
6 79067
7 91087
8 102550
9 113434
10 123725
11 133356
12 142348
13 150715
14 158455
15 165599
16 172193
17 178244
18 183797
19 188866
20 193498
21 197716
22 201582
23 205076
24 208273
25 211171
26 213813
27 216205
28 218382
29 220358
30 222149
31 223785
32 225268
33 226618
34 227847
35 228963
36 229976
37 230891
38 231734
39 232510
40 233223
41 233871
42 234456
43 234993
44 235489
45 235946
46 236370
47 236764
48 237137
49 237484
50 237807
51 238107
52 238390
53 238652
54 238898
55 239124
56 239333
57 239531
58 239720
59 239899
60 240070
61 240233
62 240389
63 240536
64 240677
65 240809
66 240932
67 241048
68 241158
69 241265
70 241366
71 241461
72 241553
73 241643
74 241725
75 241802
76 241874
77 241941
78 242006
79 242065
80 242123
81 242179
82 242234
83 242287
84 242338
85 242388
86 242436
87 242481
88 242524
89 242564
90 242603
91 242639
92 242673
93 242706
94 242739
95 242771
96 242800
97 242829
98 242855
99 242878
100 242895
101 242909
102 242919
103 242926
104 242932
105 242938
106 242944
107 242948
108 242950
109 242952
110 242953
111 242954
112 242955
};
\addlegendentry{\tiny{WaveNet}}
\end{axis}

\end{tikzpicture}}
    \caption{Cumulative distribution of matching distance for ORB feature extraction in the Cave dataset. Higher lines indicate the enhancement, on average, allows frames to match with frames further ahead in the video than lower lines.}
    \label{fig:ahead_cave}
\end{figure}
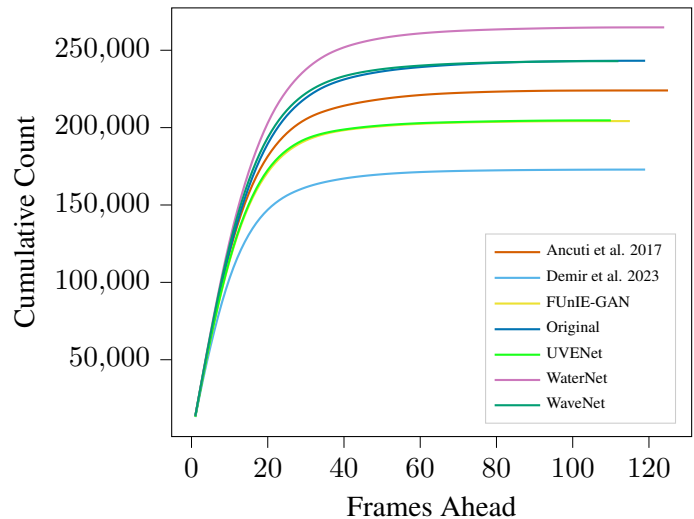

\begin{figure}[h]
    \centering
    \resizebox{0.6\textwidth}{!}{\input{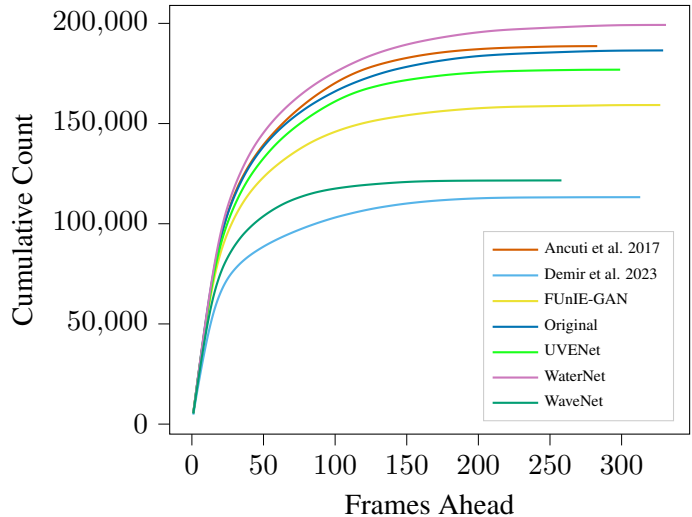}}
    \caption{Cumulative distribution of matching distance for ORB feature extraction in the Seabed dataset. Higher lines indicate the enhancement, on average, allows frames to match with frames further ahead in the video than lower lines.}
    \label{fig:ahead_seabed}
\end{figure}

Figures \ref{fig:ahead_cave} and \ref{fig:ahead_seabed} underscore much of our previous findings. \citeauthor{demir2023low} starts to plateau at the lowest values for frames ahead for both datasets, demonstrating an inability to facilitate significantly far matches compared to all other methods. \citeauthor{ancuti2017color}, UVENet, and WaveNet demonstrate inconsistent results between the datasets. Importantly, WaterNet again performs consistently well on both datasets and is the only method to show a marked improvement on the original video frames.

\subsection{ORB-SLAM Findings}

Finally we tested the impact of enhanced videos against the original in a full SLAM pipeline. We ran each enhanced or original video from both datasets through ORB-SLAM 3 and recorded tracking status, camera trajectory, loop closure logs, and 3D point cloud data. We found that although ORB-SLAM 3 was able to maintain good tracking for most of the original video, all enhanced videos resulted in significant tracking loss with poor or no recovery once tracking was lost. As a result, each enhanced video showed evidence of trajectory drift and no video, except the original, established any loop closures. Comparing the number of points mapped with the number of frames successfully tracked showed a more even result, with \citeauthor{ancuti2017color} doing the best in both datasets, and the original video still doing better than many of the enhanced videos, as well as having the more complete final 3D cloud. These results are presented in the supplementary material.

\section{Discussion and Conclusion}
\label{sec:conclusion}
Although SSIM and PSNR scores indicate that the selected enhancement methods perform without worsening the quality of the original test videos, our findings reveal that the impact of these enhancements are not positive. Testing six distinct approaches, including classical and deep learning-based methods, we found a wide range of frame matching abilities. In the majority of cases the videos produced after enhancement were equal or worse in performance compared to the unaltered original video. \citep{demir2023low} stood out as performing the worst in our evaluation, while WaterNet \citep{li2019underwater} was the best and the only method that reliably facilitated better frame matching results, although only marginally. All enhancement methods tested were detrimental to SLAM’s tracking accuracy, and loop closure detection, leading to drift and incomplete map generation. 

Evaluating enhancement performance using full SLAM is computationally prohibitive, particularly for large-scale video data due to significant memory demands. Our central insight is that \textit{frame matching} --- a core component of SLAM --- can serve as an effective surrogate metric. This approach offers substantial computational efficiency while still yielding reliable and robust results for cross-method comparison. 

Our experiments indicate strong correspondence between frame-matching metrics and the visual trajectories produced by full SLAM, yet the computational overhead is markedly reduced. Consequently, our evaluation framework provides a lightweight, scalable solution for benchmarking underwater image enhancement methods in the context of downstream applications. It offers a practical mechanism to assess whether new enhancement techniques can meaningfully support tasks such as SLAM-based localisation and navigation.

\appendix



\section*{Acknowledgments}
This research was funded by EPSRC grant number EP/S021892/1. For the purpose of open access the authors have applied a Creative Commons Attribution (CC BY) licence to any Author Accepted Manuscript version arising from this submission. We gratefully acknowledge the contributions of Catherine Seale and the industrial support from Vaarst/Beam. Their insights and involvement were instrumental in shaping the direction and development of this work, and we remain deeply appreciative of their collaboration.

\bibliographystyle{unsrtnat}
\bibliography{referenceswithdoi}

\newpage
\eject

\section{Appendix}
\label{sec:appendix}
\subsection{Additional Data}
Tables \ref{tab:inlier_cave} and \ref{tab:inlier_seabed} provide specific values for Figures \ref{fig:inlier_cave} and \ref{fig:inlier_seabed}.

\begin{table}[h]
\centering
\scriptsize 
\resizebox{\columnwidth}{!}{ 
\begin{tabular}{lcccccccccc}
\toprule
 & \multicolumn{10}{c}{\textbf{Average number of inliers for frame offset:}} \\
\cmidrule(lr){2-11}
\textbf{Enhancement Method} & \textbf{1} & \textbf{2} & \textbf{3} & \textbf{4} & \textbf{5} & \textbf{6} & \textbf{7} & \textbf{8} & \textbf{9} & \textbf{10} \\
\midrule
Ancuti et al. 2017 & 481.55 & 434.28 & 393.00 & 355.03 & 319.56 & 287.79 & 259.67 & 233.78 & 211.19 & 190.07 \\
Demir et al. 2023 & 389.42 & 344.26 & 308.34 & 276.01 & 245.87 & 219.35 & 195.76 & 173.81 & 154.44 & 136.74 \\
FUnIE-GAN & 424.67 & 378.53 & 339.95 & 304.80 & 274.28 & 245.92 & 221.17 & 198.70 & 178.44 & 160.05 \\
Original & 495.88 & 450.94 & 408.73 & 369.22 & 333.09 & 300.13 & 271.48 & 245.50 & 221.57 & 199.82 \\
WaterNet & 517.97 & \textbf{470.54} & \textbf{427.50} & \textbf{386.80} & \textbf{349.24} & \textbf{315.69} & \textbf{285.87} & \textbf{259.46} & \textbf{235.03} & \textbf{213.18} \\
WaveNet & \textbf{530.53} & 465.93 & 415.48 & 373.71 & 335.45 & 302.53 & 273.40 & 246.36 & 222.63 & 201.85 \\
\bottomrule
\end{tabular}
}
\caption{Average number of inliers for different frame offsets across various enhancement methods for the Cave video.}
\label{tab:inlier_cave}
\end{table}

\begin{table}[h]
\centering
\scriptsize 
\resizebox{\columnwidth}{!}{ 
\begin{tabular}{lcccccccccc}
\toprule
 & \multicolumn{10}{c}{\textbf{Average number of inliers for frame offset:}} \\
\cmidrule(lr){2-11}
\textbf{Enhancement Method} & \textbf{1} & \textbf{2} & \textbf{3} & \textbf{4} & \textbf{5} & \textbf{6} & \textbf{7} & \textbf{8} & \textbf{9} & \textbf{10} \\
\midrule
Ancuti et al. 2017 & 406.48 & 370.15 & 348.37 & 330.85 & 313.80 & 297.83 & 281.35 & 266.19 & 250.64 & 235.75 \\
Demir et al. 2023 & 328.03 & 294.69 & 275.08 & 257.64 & 240.95 & 225.57 & 210.74 & 195.76 & 180.22 & 166.76 \\
FUnIE-GAN & 369.97 & 340.06 & 321.03 & 304.23 & 287.30 & 270.58 & 254.42 & 238.81 & 222.39 & 206.70 \\
Original & \textbf{442.21} & \textbf{406.62} & \textbf{383.94} & \textbf{364.30} & 344.26 & 325.53 & 307.27 & 289.50 & 271.79 & 254.33 \\
WaterNet & 440.29 & 405.84 & 383.18 & 364.23 & \textbf{344.62} & \textbf{325.70} & \textbf{307.73} & \textbf{289.87} & \textbf{272.23} & \textbf{255.46} \\
WaveNet & 399.12 & 354.27 & 326.05 & 303.41 & 282.19 & 262.95 & 244.83 & 226.72 & 208.81 & 192.47 \\
\bottomrule
\end{tabular}
}
\caption{Average number of inliers for different frame offsets across various enhancement methods for the Seabed video.}
\label{tab:inlier_seabed}
\end{table}

For example, 495.88 for the original unenhanced video frames in column 1 of Table \ref{tab:inlier_cave} means that on average there will be 495.88 inliers found between frame $n$ and frame $n+1$ throughout the whole (original unenhanced) video. As the offset increases from one to ten, fewer inliers will be found due to camera movement and marine snow and other artifacts in the case of underwater images.

This is a useful and vital measure, because at many hours long and 30 frames per second of high definition footage, SLAM on underwater surveys is highly computationally expensive, and therefore, a way to accelerate the method is to reduce the number of frames. This experiement demonstrates that even three frames per second reduces accuracy and ability for SLAM to track features.

\subsection{ORB-SLAM 3 Results Extended}
\subsubsection{Motivation}
SLAM on underwater imagery is challenging due to the poor visual quality. Therefore, it would seem natural to use enhancement methods on the video frames to improve videos prior to SLAM processing. This was the motivation for this work where SLAM performance is analysed in this section. But it was found that enhancement methods degraded the performance of SLAM over using the original video. There are no available ground truth videos with positional data in an underwater ocean environment (i.e., underwater video with ground truth position, e.g., collected from GPS, in open sea conditions). Therefore, surrogates need to be used to analyse SLAM performance, along with qualitative assessment of paths. We also developed the idea of using feature matching (as the key component of SLAM) as a quantitative measure of enhancement performance, and thus this appears in the main part of the paper where we can rely on quantitative results rather than the qualitative discussion in this section.

\subsubsection{Tracking Ability}
We first compared the SLAM tracking ability in the original video with the videos produced from the frames enhanced by each method. We took the average tracking status across five runs of the original and each enhanced video. The tracking status is one of the pieces of data that ORB-SLAM 3 produces per frame, which we log during its operation and belongs to one of three statuses, initialising tracking, OK tracking, and lost/no tracking. The most consistent to register good tracking was the unaltered original video. For both the cave and seabed videos, all three deep learning-based enhancement models managed to regain tracking, while the classical methods, on average, did not regain tracking. We saw variation in the number of frames needed to initialise. For the cave video, Ancuti et al.'s enhancements proved to be the fastest at initialising, taking an average of $59$ frames to start tracking, compared to an average of $370.6$ frames for all other cave videos. However, in the case of the seabed video, the classical methods, including Ancuti et al. take the longest by far to initialise.


\begin{table}[h]
\centering
\begin{tabular}{lcc}
\toprule
 & \multicolumn{2}{c}{Frames to initialise} \\
\cmidrule(lr){2-3}
Enhancement method & Cave (\(\downarrow\)) & Seabed (\(\downarrow\))\\
\midrule
Original video & 365 & \textbf{58} \\
FUnIE-GAN \cite{islam2020fast} & 381 & 98 \\
Ancuti et al. 2017 \cite{ancuti2017color} & \textbf{59} & 270 \\
Demir et al. 2023 \cite{demir2023low} & 380 & 235 \\
Waternet \cite{li2019underwater} & 357 & 89 \\
WaveNet \cite{sharma2023wavelength} & 370 & 74 \\
\bottomrule
\end{tabular}
\caption{The average number of frames to initiate tracking.}
\label{tab:init}
\end{table}

The poor performance on the cave video prompted us to test how well each enhanced video would be initialised and tracked within a problematic area indicated by where all enhanced videos lost tracking. We performed this test starting from frame $3500$. Despite this area being problematic for SLAM when in the context of the whole video, starting from this region resulted in improved initialisation and better tracking from all enhancement methods. However, the original video remained the only video that resulted in a complete survey, suggesting the enhancements still degraded the performance of SLAM tracking overall.

\subsubsection{Trajectory and Loop Closures}

\begin{figure}[ht]
\centering
    \includegraphics[width=0.45\textwidth]{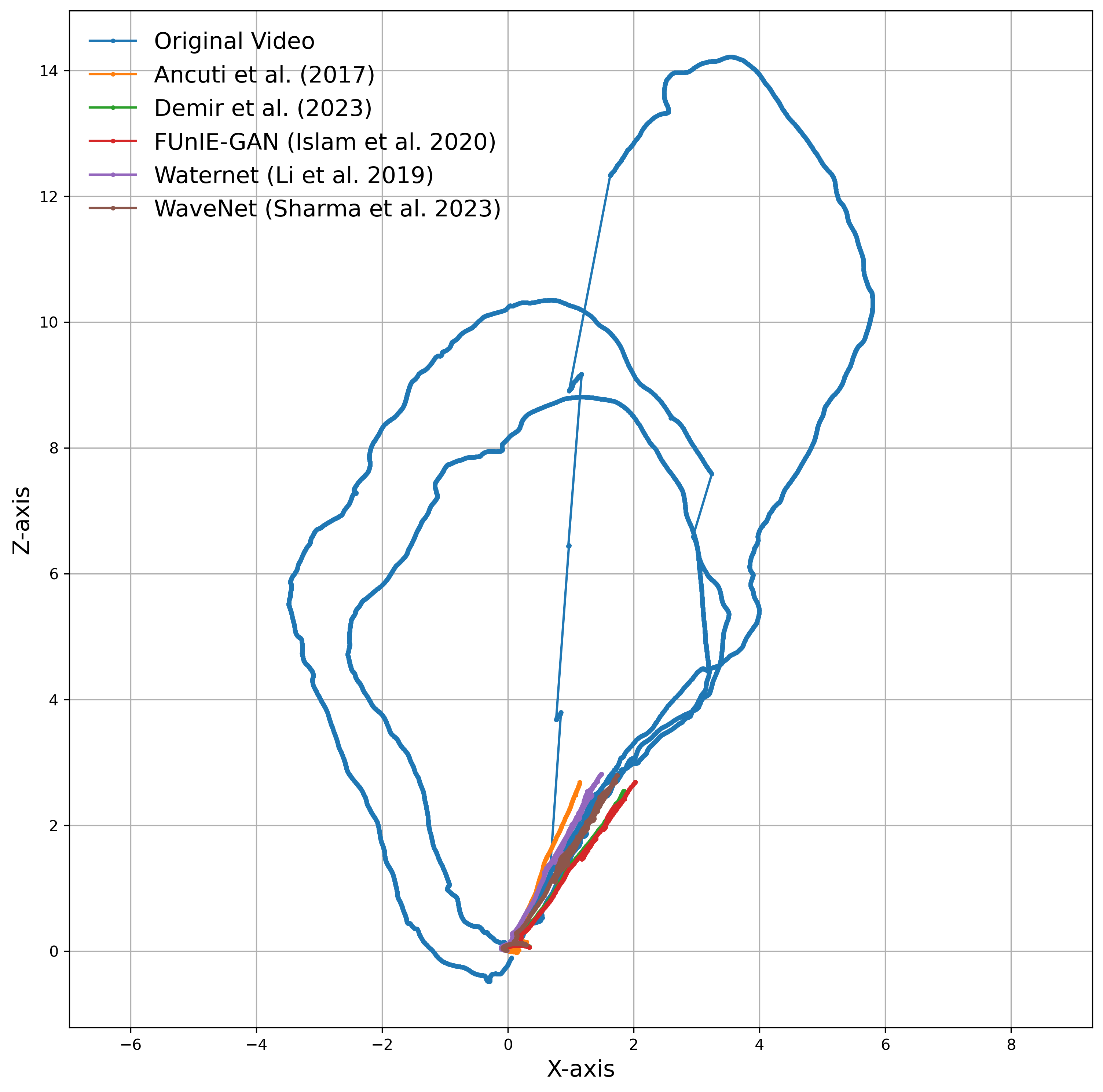}
    \caption{A top-down view of the average path for `Cave' video.}
    \label{fig:cave_path}
\end{figure}

\begin{figure}[ht]
\centering
    \includegraphics[width=0.45\textwidth]{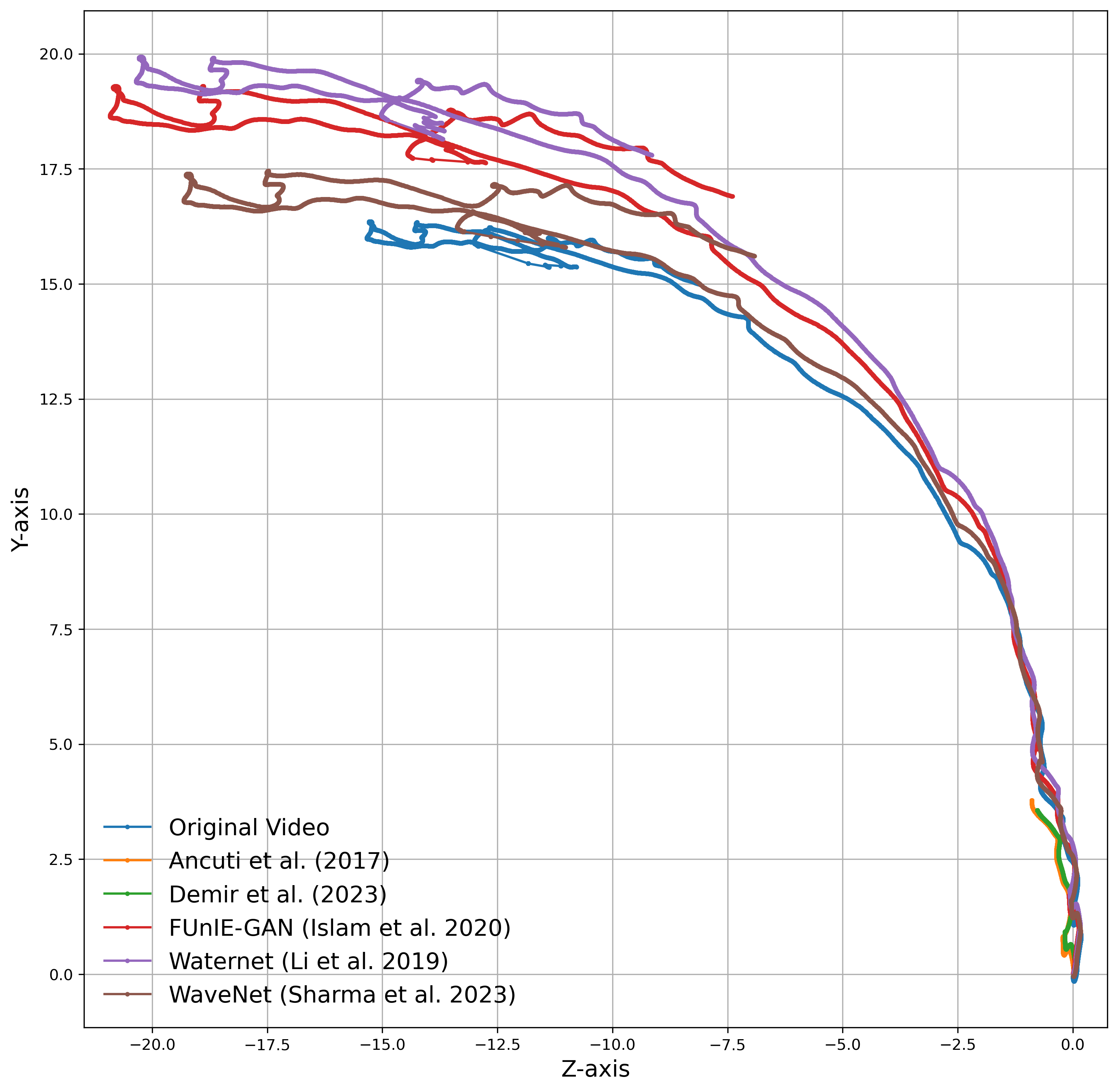}
    \caption{A side view of the average path for `Seabed' video.}
    \label{fig:seabed_path}
\end{figure}

The poor tracking performances for the enhanced videos are made more apparent by considering the recorded trajectories from SLAM. Figures \ref{fig:cave_path} and \ref{fig:seabed_path} show the average camera trajectory created from each video over five runs.

It is again clear that the original video results in the more complete SLAM survey, showing a clear, sustained path that largely matches the path visually seen in the video. A ground truth is unavailable for comparison as creating a ground truth for underwater imagery is extremely problematic, therefore, expert judgment on the path derived from SLAM, compared to the motion observed in the video is the best qualitative measure that we have. In the case of the enhanced cave videos, we see that no sustained paths were found by SLAM, and therefore there is a substantial difference between the original video and enhanced videos. More interestingly we find that in the case of the seabed video, where tracking was better, there are sustained paths from all deep-learning enhancements but with clear divergences between them. Given the encouraging tracking performance for the original video, the drift in the enhanced video surveys suggests again that the enhancements are detrimental to SLAM.

\begin{figure}[h]
\centering
    \includegraphics[width=0.45\textwidth]{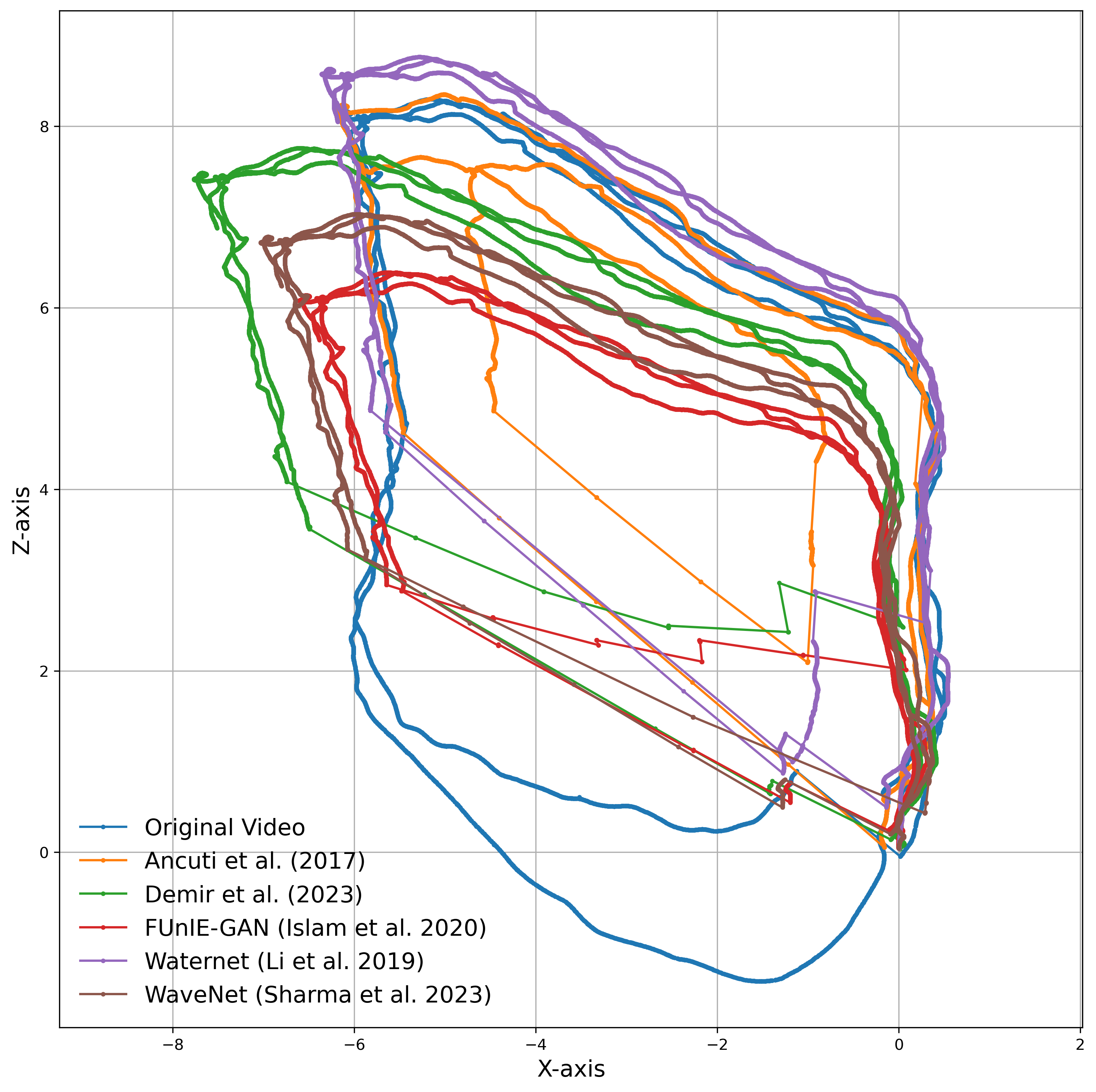}
    \caption{A top-down view of the average path for the `Cave' video started at frame 3500.}
    \label{fig:cave_3500_path}
\end{figure}

Figure \ref{fig:cave_3500_path} shows the recorded camera trajectory when SLAM is initiated in the previously problematic region of the video at frame $3500$. The result is a set of more sustained trajectories from all enhancement methods, but again there is strong evidence of drift. The figure shows that all enhancement methods resulted in half a loop before tracking was lost and then regained at the familiar beginning of the loop, preventing a full loop closure. Reinforcing this analysis are the loop closure event logs created by ORB-SLAM 3.

We found that no enhancement method resulted in the capture of a single loop closure in five runs. Only the original video resulted in any loop closure detection, with one event being identified in three of five runs, and in all five when the video started at frame $3500$. Despite being unable to complete a loop closure, Figure \ref{fig:cave_3500_path} shows that ORB-SLAM 3 was able to relocalise itself after losing track. This suggests that these enhanced videos may be capable of facilitating loop closures but as the enhancements lead to inconsistent tracking the likelihood of loop closure detection is reduced.

\end{document}